\documentclass{ecai} 

\usepackage{latexsym}
\usepackage{amssymb}
\usepackage{amsmath}
\usepackage{amsthm}
\usepackage{booktabs}
\usepackage{enumitem}
\usepackage{graphicx}
\usepackage{color}
\usepackage{multirow}
\usepackage{xcolor}
\usepackage{adjustbox}

\usepackage[normalem]{ulem}
\definecolor{darkgreen}{rgb}{0,.4,0}
\definecolor{darkcyan}{rgb}{0,.4,.4}
\newcommand{\REMOVE}[1]%
          {{\color{blue}\sout{#1}}}

\newcommand{\COMMENT}[1]%
          {{\color{darkgreen}\textbf{{Editor: }} {#1}}}

\newcommand{\BibTeX}{B\kern-.05em{\sc i\kern-.025em b}\kern-.08em\TeX}

\begin{document}


\begin{frontmatter}


\paperid{123} 


\title{Multi-dimension Transformer with Attention-based Filtering for Medical Image Segmentation}


\author[A]{\fnms{Wentao}~\snm{Wang}}
\author[A]{\fnms{Xi}~\snm{Xiao}}
\author[B]{\fnms{Mingjie}~\snm{Liu}\thanks{Corresponding Author. Email: liumj@cqupt.edu.cn}}
\author[A]{\fnms{Qing}~\snm{Tian}}
\author[B]{\fnms{Xuanyao}~\snm{Huang}}
\author[A]{\fnms{Qizhen}~\snm{Lan}}
\author[C]{\fnms{Swalpa Kumar}~\snm{Roy}}
\author[A]{\fnms{Tianyang}~\snm{Wang}}

\address[A]{University of Alabama at Birmingham}

\address[B]{Chongqing University of Posts and Telecommunications}

\address[C]{Alipurduar Government Engineering and Management College}


\begin{abstract}
The accurate segmentation of medical images is crucial for diagnosing and treating diseases. Recent studies demonstrate that vision transformer-based methods have significantly improved performance in medical image segmentation, primarily due to their superior ability to establish global relationships among features and adaptability to various inputs. However, these methods struggle with the low signal-to-noise ratio inherent to medical images. Additionally, the effective utilization of channel and spatial information, which are essential for medical image segmentation, is limited by the representation capacity of self-attention. To address these challenges, we propose a multi-dimension transformer with attention-based filtering (MDT-AF), which redesigns the patch embedding and self-attention mechanism for medical image segmentation. MDT-AF incorporates an attention-based feature filtering mechanism into the patch embedding blocks and employs a coarse-to-fine process to mitigate the impact of low signal-to-noise ratio. To better capture complex structures in medical images, MDT-AF extends the self-attention mechanism to incorporate spatial and channel dimensions, enriching feature representation. Moreover, we introduce an interaction mechanism to improve the feature aggregation between spatial and channel dimensions. Experimental results on three public medical image segmentation benchmarks show that MDT-AF achieves state-of-the-art (SOTA) performance.
\end{abstract}

\end{frontmatter}


\section{Introduction}

Segmentation plays a pivotal role in medical imaging analysis by identifying and outlining regions of interest within medical images~\cite{SHAMSHAD2023102802,10.1145/3615862}. However, relying on experts for diagnosis is time-consuming and prone to observer bias by clinical experience~\cite{10098158,degrave2023auditing}. Therefore, the automated segmentation technique is an excellent assistance tool. Convolutional neural networks (CNNs) have become the standard in computer vision and are widely adopted for medical image segmentation~\cite{YU202192}. UNet~\cite{UNet} and its variants are widely used in medical image segmentation, such as Attention UNet~\cite{Attentionu-net}, UNet++\cite{8932614}, and UNet 3+\cite{9053405}. However, due to the limitations of the convolution operation, CNN-based methods struggle to model long-range dependencies effectively, which can result in limited performance in medical image segmentation~\cite{YUAN2023109228}. Firstly, medical images often necessitate the modeling of global information to achieve reliable segmentation, as the shape and size of the region of interest can vary significantly. Secondly, CNN-based methods lack the flexibility to accommodate inputs with diverse characteristics, thereby limiting their ability to adapt to varying content within input images. Recently, vision transformers~\cite{dosovitskiy} have achieved remarkable success in various vision tasks~\cite{Marin_2023_WACV,Yu_2023_WACV,10054166}, demonstrating competitive performance compared to other CNN-based methods. Given their capability for long-range information interaction and dynamic feature encoding~\cite{10088164}, researchers have explored the use of transformers in medical image segmentation~\cite{SHAMSHAD2023102802}. TransUNet~\cite{transunet} represents the pioneering effort to integrate Transformers into medical image segmentation, utilizing a hybrid structure of CNNs and Transformers. Subsequently, a pure Transformer encoder-decoder architecture named SwinUNet~\cite{Swin-Unet} was introduced, demonstrating strong performance in this domain.

While Transformers perform well in modeling long-range dependencies, they still face drawbacks and limitations when applied to medical images. i) these methods struggle to handle the inherently low signal-to-noise ratio of medical images~\cite{GerlStefan,kaur2023complete}, which can significantly hinder feature learning and discrimination. ii) their multi-dimension representation capability is still limited, particularly in capturing channel and spatial information, which are crucial for medical image segmentation. Therefore, there remains significant room for improvement in Transformer-based methods.

To address the above limitations, we propose MDT-AF, a novel transformer variant designed for robust and precise medical image segmentation. MDT-AF comprises two primary components: Patch embedding with attention-based filtering and self-attention that extends across spatial and channel dimensions. Specifically, to address the first problem, we redesigned the patch embedding mechanism of the vision transformer by incorporating an attention-based filtering mechanism parallel to the patch embedding module. This filtering mechanism aims to refine coarse features and reduce noise, thereby enhancing the model's ability to focus on relevant signals and adapt dynamically to varying noise levels across different image regions. As a result, this mechanism improves the quality of extracted features and enhances the model's accuracy in identifying and delineating critical areas in medical images. To tackle the second challenge, we introduced multi-dimension transformer blocks, which expand self-attention to spatial and channel dimensions and perform feature interaction and aggregation within blocks. Spatial self-attention enriches the spatial expression of each feature map, while channel self-attention facilitates global information exchange between features. As a result, this mechanism improves the richness of feature representation and enhances the model's ability to model medical images with various shapes and scenes effectively.

\begin{figure*}[t]
\centering
\includegraphics[width=17cm]{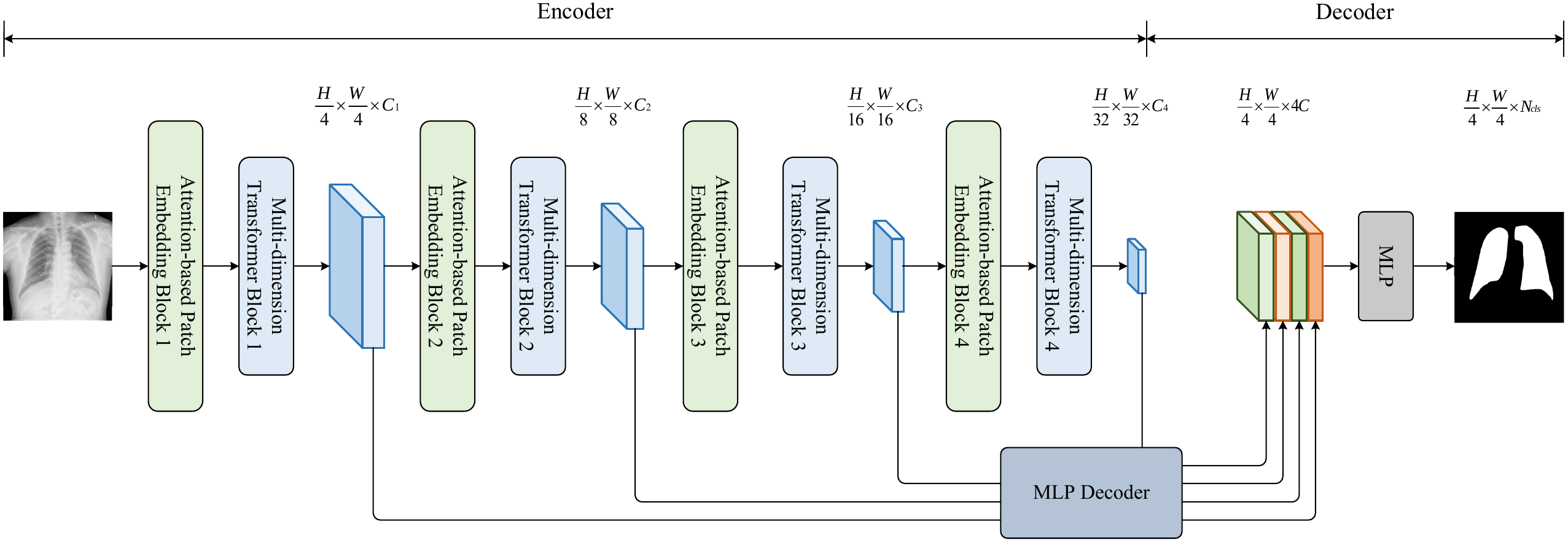}
\caption{\textbf{The proposed MDT-AF framework} comprises three primary modules: 1) Attention-based patch embedding blocks that generate patch tokens while concurrently producing attention weights. These weights are instrumental in filtering out coarse features and noise. 2) Multi-dimension transformer blocks extend self-attention across spatial and channel-wise dimensions to build and aggregate a comprehensive feature representation. 3) MLP decoders fuse these multi-level features to accurately predict the semantic segmentation mask. Where $C_{1}$ is set to 64, $C_{2}$ is set to 128, $C_{3}$ is set to 320, and $C_{4}$ is set to 512.}
\label{framework}
\end{figure*}

In contrast to existing approaches, our method offers the following three main contributions:

1) We propose a novel transformer variant for stable and precise medical image segmentation, which can generate rich feature representation while minimizing redundant information.

2) We introduce an attention-based filtering mechanism. This mechanism filters the coarse features and noise acquired from patch embedding, which can enable the construction of more refined feature representations. Moreover, the self-attention mechanism is expanded to spatial and channel dimensions and incorporates feature interaction and aggregation within blocks to establish comprehensive feature representation.

3) Experimental evaluations conducted on three public medical image segmentation benchmarks across Lung, Skin, and Polyp demonstrate the effectiveness and superiority of our method compared to the state-of-the-art methods.

\section{Related Works}

\subsection{CNN-based segmentation methods}

CNNs have been primarily used for medical image segmentation in recent years due to their powerful feature representation capabilities~\cite{Sarvamangala}. U-shaped architectures play a significant role and are widely used in medical image segmentation due to their encoder-decoder architecture~\cite{LIU2020244}. UNet, an encoder-decoder architecture featuring multiscale skip connections, has demonstrated state-of-the-art performance across various medical image segmentation tasks. Subsequently, several variants of UNet have been developed, including UNet++, Attention UNet, R2U-Net~\cite{alom2018recurrent}, and UNet3+. UNet++ utilized the advantage of dense skip connections to link its nested encoder-decoder subnetworks. The Attention UNet introduced an attention mechanism to refine the output features of the encoder before merging these features with the corresponding decoder features at each resolution level. R2U-Net is an extension of standard U-Net, which combines recurrent neural networks and residual skip connections. UNet3+ extends the concept of full-scale skip connections by incorporating intra-connections between the decoder blocks. Inspired by these architectures, specific methods have been tailored for particular tasks, such as polyp segmentation~\cite{YEUNG2021104815,9469419}, skin lesion segmentation~\cite{SUN2023109524,Jiacheng}, etc. Although these methods have improved the abilities of context modeling to some extent, CNN's limited receptive fields strand their performance.

\subsection{Transformer-based segmentation methods}

Recently, vision transformer-based methods have achieved state-of-the-art performance in various vision tasks. Given their success, numerous studies have explored the application of transformers in medical image segmentation. Compared to CNNs, transformers can capture long-range dependencies through sequence modeling and multi-head self-attention, leading to improved segmentation performance. TransUNet~\cite{transunet} integrates UNet with Transformers to learn both local and global pixel relations. TransFuse~\cite{transfuse} combines Transformers and CNNs in parallel to capture global dependencies and low-level spatial details in a more efficient and shallower manner. MedT~\cite{MedT} leverages both local and global information by employing two branches: A gated axial Transformer to explore global information, and a CNN to learn local information. SwinUNet~\cite{Swin-Unet} introduces a pure U-shaped transformer architecture using Swin transformer blocks, where all convolutional layers in the U-Net are replaced by Swin transformer blocks. DS-TransUnet~\cite{9785614} further extends SwinUNet by introducing a fusion module designed to model long-range dependencies between features of different scales. MISSformer~\cite{9994763} implements an encoder-decoder architecture using enhanced transformer blocks and introduces a convolutional layer within the Transformer to enhance its ability to capture local information. Hiformer~\cite{Heidari_2023_WACV} proposes a combination of a Swin Transformer module and a CNN-based encoder to generate two multi-scale feature representations, which are integrated via a Double-Level Fusion module.

These transformer-based methods had several shortcomings despite their encouraging results. 1) They overlook important aspects of medical image segmentation, such as the representation and aggregation of features across spatial and channel dimensions. 2) The poor signal-to-noise ratio in medical images makes these methods challenging to use and would negatively impact segmentation performance. In this study, we redesign the patch embedding and self-attention mechanism, where the modified patch embedding process provides an attention-based filtering mechanism. By generating a coarse-to-fine process, a finer representation of the features is obtained. Furthermore, we expand self-attention across channel and spatial dimensions, enabling comprehensive representation and aggregation of cross-dimension information within blocks. Finally, a novel transformer variant, termed MDT-AF, is proposed with superior performance, which can provide rich feature representation and redundant information filtering for medical image segmentation.

\section{Methods}

In this section, we first present the overview of our proposed MDT-AF. Then, the details of the proposed patch embedding with attention-based filtering and multi-dimension transformer block will be given.

\subsection{Overall Architecture}

The overall architecture of the proposed MDT-AF mainly comprises three modules: Attention-based patch embedding blocks, multi-dimension transformer blocks, and MLP layers, as illustrated in Fig~\ref{framework}. Specifically, given an input image, it is initially processed through the patch embedding block. This block comprises two parallel branches: One is the overlap patch embedding, utilized for extracting initial features, while another parallel branch is the attention-based filtering mechanism, which generates attention weights to filter the output of the first branch. Subsequently, we obtain more refined features after this filtering process, which are further passed into the transformer block. Within the transformer block, self-attention is extended across spatial and channel dimensions, enabling the model to establish and aggregate features across multi-dimension. The input image undergoes processing through four levels of patch embedding blocks and transformer blocks. The processed features are fed into the MLP decoders, which integrate features from four levels to predict semantic segmentation masks.

\begin{figure*}[t]
\centering
\includegraphics[width=16cm]{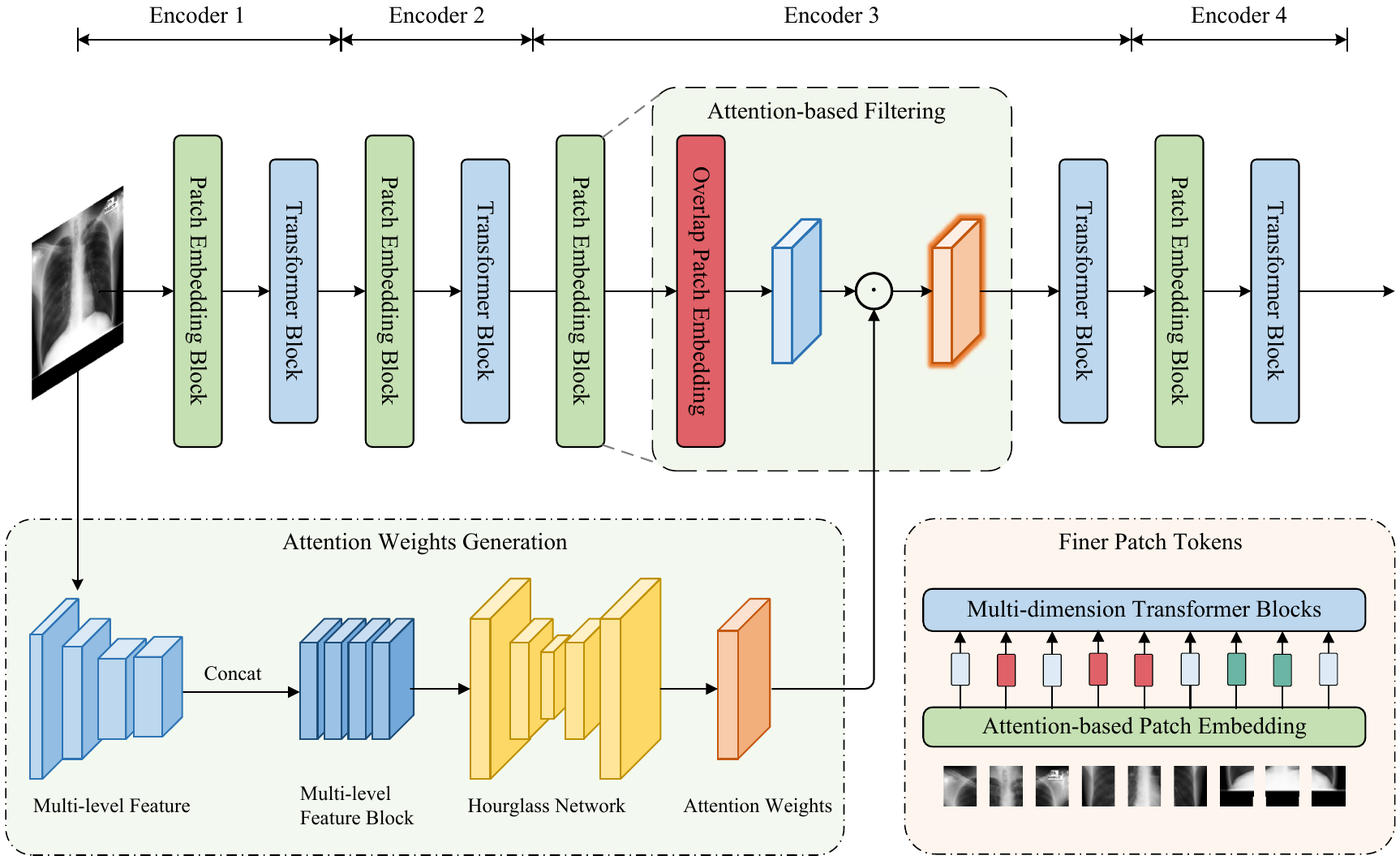}
\caption{\textbf{The proposed Patch Embedding with Attention-based Filtering} consists of two parallel branches: 1) The Overlap Patch Embedding branch processes an input feature map to extract coarse features. 2) Simultaneously, a parallel branch generates corresponding attention weights. These weights are applied to the coarse features, filtering out noise and refining the feature representation from coarse to fine. Notably, this approach is consistently employed in the patch embedding block of each encoder stage to generate patch tokens.}
\label{methods-1}
\end{figure*}

\subsection{Patch Embedding with Attention-based Filtering}

Medical images generally confront challenges related to low signal-to-noise ratio, which directly result in the blurring of critical features such as boundaries, thus complicating the pixel-level classification. Patch Embedding is a crucial component of transformer-based methods, which plays a vital role in preprocessing input images and generating embeddings. These embeddings are then processed into tokens utilized by subsequent transformer blocks for feature extraction and classification tasks. Improving Patch Embedding is essential for several reasons: i) It will provide a more robust initial feature representation, crucial for handling noisy inputs in medical images; ii) It will preserve more detailed spatial information, which is vital for identifying subtle anatomical structures in blurred images; iii) It will improve the model's efficiency by ensuring that higher-quality embeddings lead to more effective subsequent processing; iv) Optimized Patch Embeddings will be able to enhance the robustness of the model against variations in image quality, a common issue in different clinical environments.

Motivated by~\cite{xu2023memoryefficient,xu2023cgi,xu2022attention,wang2024selectivestereo}, we propose a patch embedding with an attention-based filtering mechanism. Specifically, we incorporate an attention-based filtering branch parallel to the patch embedding module to filter the coarse features and noise, as is shown in Fig~\ref{methods-1}. This mechanism aims to refine the feature representation by selectively enhancing relevant features and suppressing noise, thereby improving the segmentation accuracy in medical images. Next, we will describe the details below.

\textit{\textbf{1) Overlap Patch Embedding:}} The initial design of patch embedding in Vision Transformers (ViT)~\cite{dosovitskiy} involved combining non-overlapping images or feature patches. However, this method can not effectively preserve local continuity around the patches. In this study, we employ overlapping patch embedding to better capture features~\cite{NEURIPS2021_64f1f27b}. In our experiments, we applied two configurations to facilitate overlapping patch embedding: $K$ = 7, $S$ = 4, $P$ = 3, and $K$ = 3, $S$ = 2, $P$ = 1. The coarse features obtained from overlap patch embedding are recorded as $F_{1}$.

\textit{\textbf{2) Attention-based Filtering:}} Given an input $X$, we first use a $1 \times 1$ convolution operation to reshape the channels of the input to 40, noted as $X_{r}$. Then, a convolution with $3 \times 3$ kernel size and 40 groups is applied to the $X_{r}$, noted as $X_{gwc}$ Next, the first 8 channels of the grouped features are further processed by a $3 \times 3$ convolution with a dilation of 1 (Level 1). The next 16 channels are processed by a $3 \times 3$ convolution with a dilation of 2 (Level 2). The last 16 channels are processed by a $3 \times 3$ convolution with a dilation of 3 (Level 3). The process is defined as:

\begin{align}
V_{l1} &= \text{DilatedConv}_{l1}(X_{\text{gwc}}[:, :8]), \\
V_{l2} &= \text{DilatedConv}_{l2}(X_{\text{gwc}}[:, 8:24]), \\
V_{l3} &= \text{DilatedConv}_{l3}(X_{\text{gwc}}[:, 24:40])
\end{align}

where $\text{DilatedConv}_{li}$ corresponds to the dilated convolution operation for level $i$ with different dilation rates and group settings. The outputs of these three levels are concatenated along the channel dimension, resulting in a combined feature map that includes multi-scale feature information, which can be formulated as:

\begin{equation}
V_{\text{mpmv}} = \text{Concat}(V_{l1}, V_{l2}, V_{l3})
\end{equation}

Subsequently, we apply two convolutions and a 2D hourglass network~\cite{Guo_2019_CVPR} to regularize $V_{\text{mpmv}}$. Following this, another convolutional layer compresses the channels to 1, deriving the attention weights denoted as $A$. With the obtained attention weights, we utilize them to eliminate redundant information in the coarse feature. The output feature $X_{out}$ at channel $i$ is computed as:

\begin{equation}
X_{out}(i)={A}\odot{F}_{1},
\label{equ:acv}
\end{equation}

where $\odot$ indicates the element-wise product, and the attention weights $A$ are applied to all channels of the coarse feature.

\subsection{Multi-dimension Transformer Block}

\begin{figure*}[t]
\centering
\includegraphics[width=17cm]{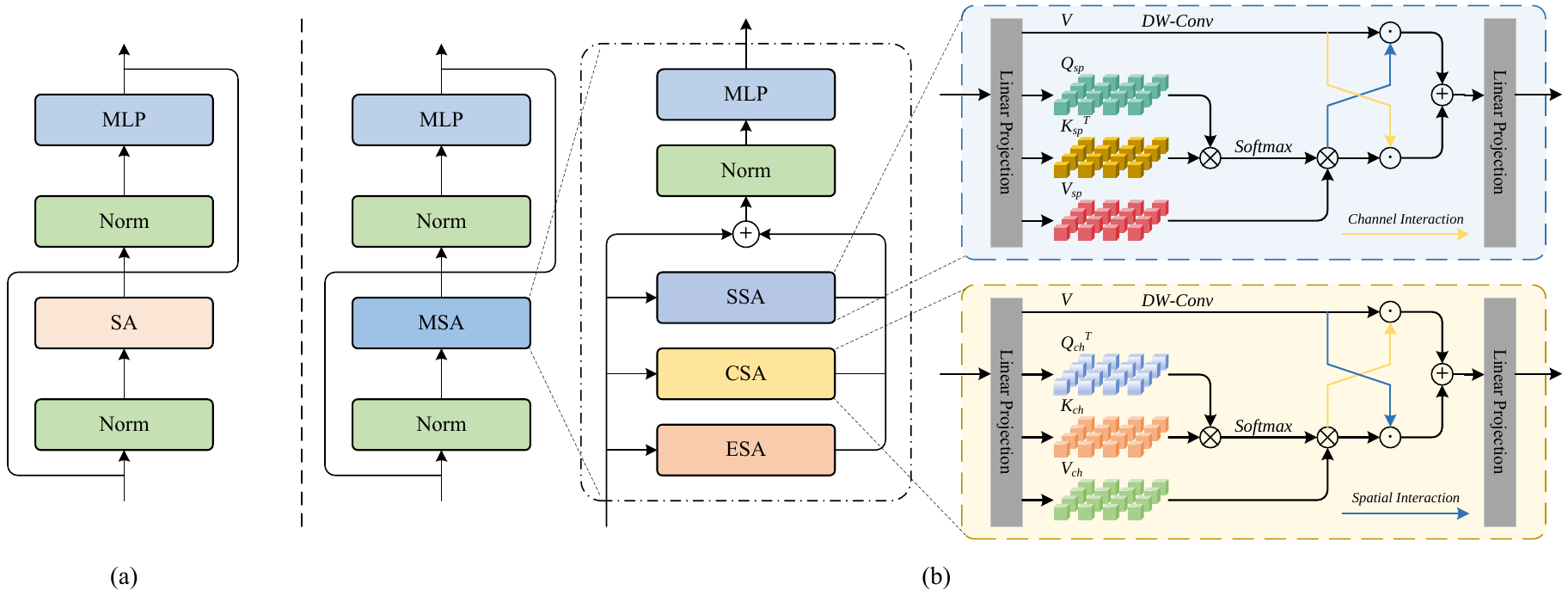}
\caption{\textbf{The proposed Multi-dimension Transformer Block (b)} expands self-attention to spatial and channel dimensions. Unlike transformer block (a), it incorporates feature interaction and aggregation within blocks, with spatial self-attention capturing contextual information across image positions and channel self-attention analyzing feature channel relationships to highlight significant features. And a convolution branch is parallel with self-attention to add locality. "SA" denotes self-attention, "ESA" signifies efficient self-attention, "SSA" stands for spatial self-attention, and "CSA" represents channel self-attention.}
\label{multi-dimension}
\end{figure*}

A rich representation of channel and spatial dimensions is essential in medical image analysis. Self-attention can not model spatial and channel information well, as it lacks spatial expression of each feature map and ignores the relationship among channels. Inspired by~\cite{li2023dlgsanet,CHEN2024104485,chen2023dual,Chen_2023_ICCV}, we introduce the Multi-dimension Transformer block during the feature extraction, mainly consisting of efficient self-attention attention, spatial dimension self-attention, and channel dimension self-attention, as shown in Fig~\ref{multi-dimension}. Specifically, spatial and channel dimension self-attentions are designed to model spatial and channel information. Considering that self-attention primarily captures global features, we integrated a convolutional branch parallel to the self-attention module to introduce locality into the Transformer architecture. And an interaction mechanism is introduced to adaptively re-weight features from the spatial or channel dimensions, allowing for better fusion of features from the two branches.

\textit{\textbf{1) Efficient Self-attention (ESA)}}: Self-attention~\cite{NIPS2017_3f5ee243} calculates the attention matrix using $queries$ ($Q$), $keys$ ($K$), and $values$ ($V$) to be a sequence of $H \times W$ feature vectors with dimensions $D$, where $H$ and $W$ is the height and width of the image, respectively. The formula is as follows:

\begin{equation}
V' = Softmax\left(\frac{Q^{T}K}{\sqrt{D}}\right)V^{T}.
\label{eqn:standard}
\end{equation}

\noindent the operation has $O(n^2)$ complexity. While for the
efficient self-attention~\cite{NEURIPS2021_64f1f27b}, a modified equation reduces computation burden:

\begin{equation}
    \begin{aligned}
    &{\hat{K}} = {\text{Reshape}(\frac{N}{R},C \cdot R)(K)}\\
    &{K} = {\text{Linear}(C \cdot R, C)({\hat{K}}),}
    \end{aligned}
    \label{eqn:reduce}
\end{equation}

where $K$ is the sequence to be reshaped, $\text{Reshape}\left(\frac{N}{R}, C \cdot R\right)(K)$ reshapes $K$ to $\frac{N}{R} \times (C \cdot R)$, and $\text{Linear}(C_{in}, C_{out})(\cdot)$ is a linear layer transforming an input tensor of dimension $C_{in}$ to an output tensor of dimension $C_{out}$. Consequently, the new $K$ has dimensions $\frac{N}{R} \times C$, resulting in reduced complexity of $O\left(\frac{N^2}{R}\right)$. In this case, given an input $X_{\text{in}}$, we get the output from Efficient Self-attention (ESA) as follows:

\begin{equation}
\label{eq:linear_projection}
Y_{E} = ESA(X_{\text{in}}) + X_{\text{in}}
\end{equation}

where $Y_{E}$ represents the output from the ESA operation.

\textit{\textbf{2) Spatial Self-attention (SSA)}}: Spatial-wise expression is a benefit for accurate boundary delineation and shape representation. To enrich the spatial expression of each feature map, we expand self-attention to the spatial dimension, as shown in Fig~\ref{multi-dimension}. Given a feature map \( f \in \mathbb{R}^{H \times W \times C} \), we first derive the query (\(Q\)), key (\(K\)), and value (\(V\)) through linear projection:

\begin{equation}
\label{eq:linear_projection}
Q = f W_{\text{Q}}, \quad K = f W_{\text{K}}, \quad V = f W_{\text{V}}
\end{equation}

where ${W_Q}, {W_K}, {W_V}$$\in$$\mathbb{R}^{C \times C}$ are the projection matrices without bias terms. We then partition ${Q}, {K}, \text{and } {V}$ into non-overlapping windows and flatten each window, noted as \( Q_{\text{sp}}, K_{\text{sp}}, \) and \( V_{\text{sp}} \), respectively. Next, we distribute them into \( h \) attention heads: \( Q_{\text{sp}} = [Q_{\text{sp}}^1, \dots, Q_{\text{sp}}^h] \), \( K_{\text{sp}} = [K_{\text{sp}}^1, \dots, K_{\text{sp}}^h] \), and \( V_{\text{sp}} = [V_{\text{sp}}^1, \dots, V_{\text{sp}}^h] \). Each head's output is then computed as follows:

\begin{equation}
\begin{gathered}
Y_{\text{sp}}^i = Softmax\left(\frac{Q_{\text{sp}}^i (K_{\text{sp}}^i)^T}{\sqrt{d}} + D\right) \cdot V_{\text{sp}}^i
\label{eq:head_output}
\end{gathered}
\end{equation}

where \( D \) signifies the relative position encoding. By reshaping and concatenating the outputs \( Y_{\text{sp}}^i \) from all heads, we acquire the feature map \( Y_{\text{sp}} \):

\begin{equation}
\begin{gathered}
Y_{Sp} = \text{concat}(Y_{\text{sp}}^1, \dots, Y_{\text{sp}}^h) W_{\text{merge}}
\label{eq:final_feature_map}
\end{gathered}
\end{equation}

where $W_{\text{merge}}$ is the linear projection to fuse all features. To better encode spatial location information and inject inductive bias, we adopt depthwise convolution as a parral branch to extract local spatial features. Specifically, given a feature map \( f \), we adopt the convolutional block to model local spatial features, which are shown as follows:

\begin{equation}
\label{eq:linear_projection}
Y_{local} = Dw\text{-}Conv(f)
\end{equation}

$Y_{local}$ is the local features, which contain 2D spatial location information, making it possible to encode position information. Given two features $X_{1}$ and \( X_{2} \in \mathbb{R}^{H \times W \times C} \), we apply two interaction mechanisms to better fuse and aggregate these features:

\begin{equation}
\begin{gathered}
I_{s}(X_{\text{1}},X_{\text{2}}) = X_{\text{1}} \odot \sigma (W_{\text{sp2}} \cdot GELU(W_{\text{sp1}} \cdot X_{\text{2}})), \\
I_{c}(X_{\text{1}},X_{\text{2}}) = X_{\text{1}} \odot \sigma (W_{\text{ch2}} \cdot GELU(W_{\text{ch1}} \cdot GAP(X_{\text{2}})))
\label{eq:sw-sa}
\end{gathered}
\end{equation}

where $\sigma$ indicates sigmoid activation. $GAP$ refers to the global average pooling layer. $W$ indicates the weight of the point-wise convolution for downscaling or upscaling channel dimensions. The ratios are set to $W_{\text{sp1}}=r_1 \downarrow, W_{\text{sp2}}=\frac{C}{r_1} \downarrow$ and $W_{\text{ch1}}=r_2 \downarrow, W_{\text{ch2}}=r_2 \uparrow$. In this case, given an input $X_{\text{in}}$, we get the output from Spatial Self-attention (SSA) as follows:

\begin{equation}
\begin{gathered}
Y_{S} = \left(I_{c}(Y_{Sp}, Y_{local}) + I_{s}(Y_{local}, Y_{Sp})\right)W_{\text{merge}} + X_{\text{in}}
\label{eq:sw-sa}
\end{gathered}
\end{equation}

where $W_{\text{merge}}$ indicates the integrating projection matrix, $Y_{S}$ represents the output from the SSA operation.

\textit{\textbf{3) Channel Self-attention (CSA)}}: Capturing intricate channel-wise relationships is crucial for medical image segmentation with different shapes and scales. Thus, we expand self-attention to the channel dimension, enhancing feature representation by focusing on inter-channel relationships, which complements token-wise self-attention in Transformer. Given an input \( f \), linear projections are utilized to generate channel queries (\( Q_{\text{ch}} \)), keys (\( K_{\text{ch}} \)), and values (\( V_{\text{ch}} \)), all are reshaped to \( \mathbb{R}^{HW \times C} \). Following a similar process above, the projection vectors are split into \( h \) heads. The channel-wise attention for the \( i \)-th head is computed as:

\begin{equation}
\begin{gathered}
Y_{\text{ch}}^i = V_{\text{ch}}^i \cdot \text{softmax}\left(\frac{(Q_{\text{ch}}^i)^T K_{\text{ch}}^i}{\alpha}\right), \\
Y_{Ch} = \text{concat}(Y_{\text{ch}}^1, \dots, Y_{\text{ch}}^h) W_{\text{merge}}
\label{eq:a-ch-sa}
\end{gathered}
\end{equation}

where \( Y_{\text{ch}}^i \) represents the output of the \( i \)-th head, enhancing channel-specific features. \( \alpha \) is a learnable scaling factor that modulates the attention mechanism's sensitivity. The feature map \( Y_{Ch} \) is obtained by concatenating and reshaping all \( Y_{\text{ch}}^i \) outputs. Similarly, in this case, given an input $X_{in}$, we get the output from Channel Self-attention (CSA) as follows:

\begin{equation}
\begin{gathered}
Y_{C} = \left(I_{s}(Y_{Ch}, Y_{local}) + I_{c}(Y_{local}, Y_{Ch})\right)W_{\text{merge}} + X_{\text{in}}
\label{eq:sw-sa}
\end{gathered}
\end{equation}

The output of the multi-dimension Transformer block can be formulated as:

\begin{equation}
Y = MLP(Y_{E} + \lambda_{1} Y_{S} + \lambda_{2}Y_{C})
\end{equation}

where $MLP$ denoted the multilayer perceptron, $\lambda_{1}$ is set to 0.6, and $\lambda_{2}$ is set to 0.4. Finally, combining spatial and channel dimension information with initial self-attention facilitates comprehensive feature representation for medical image segmentation.

\begin{table*}[t]
\caption{Comparison with state-of-the-art methods was conducted on the Lung X-ray, Skin Lesion, and Kvasir-SEG datasets. The evaluation highlighted the best-performing methods with bold for the highest dice score (DSC) and accuracy (Acc). The second-best results were marked in {\color{red}{red}}, while the third-best results were indicated in {\color{blue}{blue}}.}
\centering
\begin{tabular}{@{}lcccccccccccccc@{}}
\toprule
\multirow{2}{*}{Model} & \multicolumn{2}{c}{Lung X-ray} & \multicolumn{2}{c}{Skin Lesion} & \multicolumn{2}{c}{Kvasir-SEG}\\ 
\cmidrule(ll){2-3} \cmidrule(ll){4-5} \cmidrule(ll){6-7}
& Acc($\%$) $\uparrow$ & DSC($\%$) $\uparrow$ & Acc($\%$) $\uparrow$ & DSC($\%$) $\uparrow$ & Acc($\%$) $\uparrow$ & DSC($\%$) $\uparrow$ \\
\midrule
UNet~\cite{UNet}  \raisebox{0.5pt}{\color{gray}\scriptsize [MICCAI'15]} &  96.67 & 94.27 & 93.86 & 83.54 & 95.62 & \color{blue}{87.96} \\

AttnUNet~\cite{Attentionu-net}  \raisebox{0.5pt}{\color{gray}\scriptsize [MIDL’18]} & \color{red}{98.11} & \color{red}{95.97} & 93.74 & 83.27 & 95.49 & 87.14 \\

DeepLabv3+~\cite{Chen_2018_ECCV}  \raisebox{0.5pt}{\color{gray}\scriptsize [ECCV’18]} & 97.54 & 94.51 & \color{red}{96.65} & \color{red}{91.03} & \color{red}{96.45} & 86.35 \\

UNet++~\cite{8932614}  \raisebox{0.5pt}{\color{gray}\scriptsize [TMI'20]} & 96.69 & 94.33 & 93.59 & 82.96 & 95.51 & 87.01 \\

UNet3+~\cite{9053405}  \raisebox{0.5pt}{\color{gray}\scriptsize [ICASSP'20]} & \color{blue}{97.99} & \color{blue}{95.63} & 95.18 & 88.52 & 95.19 & 83.95 \\

TransUNet~\cite{transunet}  \raisebox{0.5pt}{\color{gray}\scriptsize [ArXiv’21]} & 97.58 & 95.25 & 93.32 & 82.25 & \color{blue}{96.40} & \color{red}{89.21}\\

MedT~\cite{MedT} \raisebox{0.5pt}{\color{gray}\scriptsize[MICCAI’21]} & 96.07 &  91.91 & 92.18 & 81.83 & 88.96 & 45.97 \\

SwinUNet~\cite{Swin-Unet}  \raisebox{0.5pt}{\color{gray}\scriptsize [ECCVW’22]} & 95.88 & 90.67 & 93.46 & 84.01 & 93.31 & 70.71 \\

Missformer~\cite{9994763}  \raisebox{0.5pt}{\color{gray}\scriptsize [TMI’22]} & 97.86 & 95.49 & 93.93 & 84.50 & 92.98 & 71.56 \\

Hiformer~\cite{Heidari_2023_WACV}  \raisebox{0.5pt}{\color{gray}\scriptsize [WACV’23]} & 97.79 & 94.73 & \color{blue}{95.57} & \color{blue}{88.59} & 94.51 & 80.51 \\

MERIT~\cite{rahman2023multi} \raisebox{0.5pt}{\color{gray}\scriptsize [MIDL’23]} & 89.85 & 73.20 & 94.47 & 79.73 & 93.94 & 76.85 \\

\midrule
MDT-AF \raisebox{0.5pt}{\color{gray}\scriptsize [Ours]} & \textbf{98.60} & \textbf{97.17} & \textbf{97.77} & \textbf{94.25} & \textbf{98.12} & \textbf{93.38} \\ 
\bottomrule
\end{tabular}
\label{Comparison}
\end{table*}

\section{Experimental setup}

\subsection{Datasets}

To comprehensively validate the effectiveness of MDT-AF, we use three widely used medical image datasets, each acquired with different imaging devices and capturing distinct subjects. The metrics we employ for the evaluation include Accuracy (Acc) and Dice Score (DSC).

\textit{\textbf{1) Lung X-ray Dataset}}: The dataset was acquired jointly by Shenzhen Hospital, China~\cite{ref27}, and the tuberculosis control program of the Department of Health and Human Services of Montgomery County, MD, USA~\cite{ref27_add1}. It comprises 704 chest X-ray (CXR) images and corresponding label masks. We conducted 5-fold cross-validation and reported the average results across these folds.

\textit{\textbf{2) Skin Lesion Dataset}}: This study presents a dermoscopic image analysis benchmark challenge aimed at automated skin cancer diagnosis. We utilized a fusion dataset comprising images from ISIC 2017~\cite{codella2018skin}, ISIC 2018~\cite{codella2019skin}, and the PH2 dataset~\cite{6610779}. Specifically, the dataset used for this study consists of 2794 samples for training and an additional 600 samples for testing purposes.

\textit{\textbf{3) Kvasir-SEG Dataset}}: The Polyps Dataset (Kvasir-SEG) is a publicly available dataset that includes 1000 polyp images along with their corresponding segmentation masks~\cite{Kvasir-SEG}. In this study, we conducted 5-fold cross-validation and reported the average results across these validation folds.

\subsection{Implementation details}

All experiments were conducted using PyTorch and trained on a single Nvidia A100 GPU. Input medical images were resized to 512×512 for consistency in comparison. We employed the AdamW optimizer with a momentum of 0.9 and weight decay of 1e-2 to train our model for 100 epochs with a batch size of 8. A cosine learning rate scheduler was utilized during training, with maximum and minimum learning rates set to 1e-4 and 1e-6, respectively. For the loss function, we only utilized CrossEntropy loss, defined as shown in Formula~\ref{CELOSS}:

\begin{equation}
\begin{gathered}
Loss = -\frac{1}{N} \sum_{i=1}^{N} \left[y_i \log(\hat{y}_i) + (1 - y_i) \log(1 - \hat{y}_i)\right]
\label{CELOSS}
\end{gathered}
\end{equation}

where $N$ represents the number of samples, $y_i$ denotes the label of the $i^{th}$ sample, and $\hat{y}_i$ indicates the predicted probability of the $i^{th}$ sample.

\section{RESULTS}
\label{RESULTS}

\subsection{Comparison with state-of-the-arts}

\begin{figure*}[t]
\centering
\includegraphics[width=18cm]{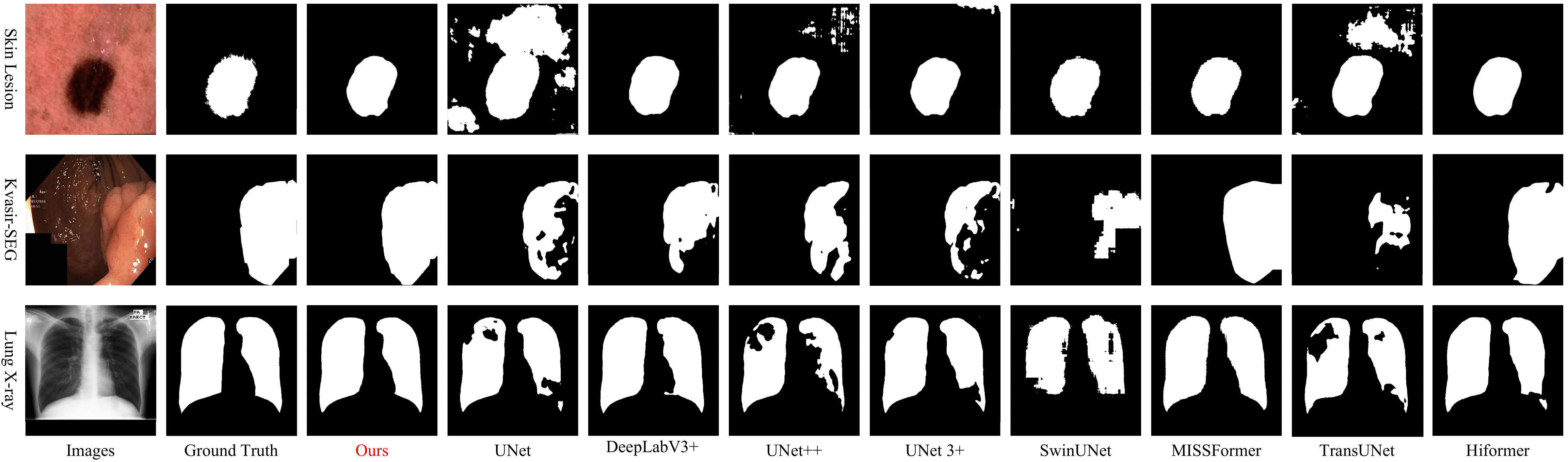}
\caption{The visual comparison results are presented for the Lung X-ray, Skin Lesion, and Kvasir-SEG datasets. The images in the first row depict the results of skin lesion segmentation, the second row shows the outcomes of polyp segmentation, and the final row displays the results of lung segmentation.}
\label{visualization}
\end{figure*}

\textit{\textbf{1) Results of Lung X-ray Dataset}}: The comparison of our MDT-AF with state-of-the-art (SOTA) methods on the Lung X-ray Dataset is presented in Table~\ref{Comparison}. Our MDT-AF achieves 98.60\% accuracy (Acc) and 97.17\% Dice Score (DSC) on this dataset. MDT-AF demonstrates significant performance improvements over CNN-based methods, i.e., AttnUNet~\cite{Attentionu-net} (+0.49\% Acc and +1.2\% DSC) and UNet3+~\cite{9053405} (+0.61\% Acc and +1.54\% DSC). Compared to transformer-based methods, MDT-AF remains competitive, surpassing four widely recognized models: TransUNet~\cite{transunet} (+1.02\% Acc and +1.92\% DSC), SwinUNet~\cite{Swin-Unet} (+2.72\% Acc and +6.5\% DSC), Missformer~\cite{9994763} (+0.74\% Acc and +1.68\% DSC), and Hiformer~\cite{Heidari_2023_WACV} (+0.81\% Acc and +2.44\% DSC). Qualitative results of some methods are depicted in Fig~\ref{visualization}. We can observe that MDT-AF accurately segments delicate and complex structures with more precise boundaries, showing robustness against complicated backgrounds. In contrast, many baseline methods like TransUNet and Hiformer struggle to locate regions of interest precisely and exhibit misclassified pixels.

\textit{\textbf{2) Results of Skin Lesion Dataset}}: To further demonstrate the generalization capability of our MDT-AF, we evaluated it on the Skin Lesion Dataset, and the experimental results are summarized in Table~\ref{Comparison}. Our MDT-AF performs better than state-of-the-art methods, achieving 97.77\% accuracy (Acc) and 94.25\% Dice Score (DSC). Figure~\ref{visualization} further compares skin lesion segmentation visual results, illustrating that our method captures finer structures and produces more precise contours. As depicted in Figure~\ref{visualization}, our method outperforms hybrid methods like Hiformer, particularly in boundary areas. Additionally, MDT-AF demonstrates greater robustness to noisy elements than purely transformer-based methods such as SwinUNet. This performance results from the superiority of our attention-based patch embedding and transformer blocks with multi-dimension self-attention.

\textit{\textbf{3) Results of Kvasir-SEG Dataset}}: In Table~\ref{Comparison}, we present results from the Kvasir-SEG dataset, where the MDT-AF architecture consistently outperforms state-of-the-art methods. Additionally, Figure~\ref{visualization} showcases segmentation outputs of the proposed MDT-AF, highlighting how our predictions closely align with the provided ground truth. MDT-AF's key advantage lies in its ability to achieve multi-dimension information representation and aggregation. Moreover, it effectively suppresses background noise through the attention-based filtering mechanism in the patch embedding process. In contrast, other methods easily encounter issues like boundary-blurring and misclassification when processing endoscopic images.

\begin{table}[t]
\caption{Ablation studies were conducted on each component using the Lung X-ray, Skin Lesion, and Kvasir-SEG datasets. The components evaluated include Patch Embedding (PE), Patch Embedding with Attention-based Filtering (AF), Transformer Blocks with Efficient Self-attention (ESA), and Transformer Blocks with Multi-dimension Self-attention (MSA). The best results are highlighted in bold.}
\label{ablation}
\centering
\begin{tabular}{@{}ccccccc@{}}
\toprule
\multicolumn{4}{c}{Module} & \multicolumn{3}{c}{DSC (\%) $\uparrow$} \\ 
\cmidrule{1-4} \cmidrule{5-7}
PE  & AF & ESA & MSA & Lung X-ray & Skin Lesion & Kvasir-SEG \\ 
\midrule
$\checkmark$ & & $\checkmark$ & & 96.84 \color{blue}{(-0.33)} & 94.16 \color{blue}{(-0.09)} & 92.27 \color{blue}{(-1.11)}\\
& $\checkmark$ & $\checkmark$ & & 97.12 \color{blue}{(-0.05)} & 94.20 \color{blue}{(-0.05)} & 92.95 \color{blue}{(-0.43)} \\
$\checkmark$ &  & & $\checkmark$ & 97.12 \color{blue}{(-0.05)} & 94.19 \color{blue}{(-0.06)} & 92.60 \color{blue}{(-0.78)} \\
& $\checkmark$ & & $\checkmark$ & \textbf{97.17} & \textbf{94.25} & \textbf{93.38} \\ 
\bottomrule
\end{tabular}
\end{table}

\subsection{Ablation study}

Extensive ablation studies were conducted on the Lung X-ray, Skin Lesion, and Kvasir-SEG datasets in order to verify the effectiveness of the attention-based filtering mechanism and transformer with multi-dimension self-attention. The Dice score (DSC) was selected as the default evaluation metric, and quantitative results are reported in Table~\ref{ablation}.

\textit{\textbf{1) Effectiveness of the Attention-based Filtering}}: In this experiment, we tested two variants of MDT-AF: a) PE+ESA: This variant removes the attention-based filtering procedure, relying solely on patch embedding to generate all patch tokens. b) AF+ESA: This variant uses a coarse-to-fine approach to generate finer patch tokens by incorporating attention-based filtering to guide the patch embedding process. The quantitative results in Table~\ref{ablation} demonstrate that both the model equipped with the attention-based filtering mechanism and the overall MDT-AF model outperform the baseline method across several datasets. Specifically, improvements over the baseline are observed in the Lung X-ray (+0.05\% and +0.33\%), Skin Lesion (+0.05\% and +0.09\%), and Kvasir-SEG datasets (+0.43\% and +1.11\%).

\textit{\textbf{2) Strength of the Multi-dimension Self-attention}}: The multi-dimension self-attention module was removed in order to create a baseline model (PE+ESA). The models that use transformer blocks with multi-dimension self-attention (PE+MSA and MDT-AF) show significant improvements over the baseline. Specifically, improvements were observed on the Lung X-ray (+0.05\% and +0.33\%), Skin Lesion (+0.06\% and +0.09\%), and Kvasir-SEG datasets (+0.78\% and +1.11\%). The model can learn and aggregate richer feature representation across spatial and channel by integrating multi-dimension self-attention mechanism, which improves overall performance and feature discriminative ability.

\section{Conclusion}

In this paper, we introduce the MDT-AF, a novel transformer variant (MDT-AF) customized for precise and robust medical image segmentation. This model combines patch embedding with an attention-based filtering mechanism to provide a coarse-to-fine process, where coarse features with noise will be filtered for a finer feature representation. Low signal-to-noise ratio is a significant challenge in medical image segmentation, and this design demonstrated its ability to solve it. Additionally, richer feature representation can be captured via the modified self-attention mechanism, which effectively constructs and aggregates multi-dimension information across the spatial and channel. It provides advantages in processing the complex structures of medical images across various scales. The effectiveness and competitiveness of the MDT-AF have been proved by its superior performance, which is higher than the state-of-the-art methods on three publicly available medical image segmentation benchmarks without any advanced training strategies. In the future, we will explore the extension of MDT-AF to other downstream medical image analysis tasks.

\begin{ack}

\end{ack}



\clearpage

\bibliography{mybibfile}

\end{document}